\documentclass[11pt]{article}

\usepackage{acl}
\usepackage[T1]{fontenc}
\usepackage[utf8]{inputenc}
\usepackage{times}
\usepackage{latexsym}
\usepackage{microtype}
\usepackage{booktabs}
\usepackage{amsmath}
\usepackage{amssymb}
\usepackage{amsfonts}
\usepackage{amsthm}
\usepackage{graphicx}
\usepackage{multirow}
\usepackage{array}
\usepackage{xcolor}
\usepackage{url}
\usepackage{float}
\usepackage{hyperref}
\hypersetup{hidelinks}
\usepackage{tikz}
\usepackage{pgfplots}
\usepackage{float}

\pgfplotsset{compat=1.18}
\usepgfplotslibrary{fillbetween}
\usetikzlibrary{arrows.meta,positioning,calc}

\theoremstyle{plain}
\newtheorem{theorem}{Theorem}

\theoremstyle{definition}

\newtheorem{remark}[theorem]{Remark}

\newcommand{\D}{\mathcal{D}}

\newcommand{\Sset}{\mathcal{S}}

\newcommand{\I}{\mathcal{I}}

\newcounter{algorithm}
\newenvironment{algorithm}[1][t]
  {\begin{figure}[#1]\refstepcounter{algorithm}\small}
  {\end{figure}}
\newcommand{\algcaption}[1]{%
  \centering\textsc{Algorithm \thealgorithm:} #1\par\smallskip
  \hrule\smallskip}

\title{CODS: Iterative Bellman-Residual Data Selection for Reusable Offline Reinforcement Learning}

\author{
\textbf{Ibne Farabi Shihab}\textsuperscript{1}\thanks{Corresponding author: \texttt{ishihab@iastate.edu}}
\quad
\textbf{Sanjeda Akter}\textsuperscript{1}
\quad
\textbf{Abu Sa-Adat Mohamed Moon-Im Al Ahsan}\textsuperscript{2}
\\
\textbf{Md Najmus Swaqeeb}\textsuperscript{2}
\quad
\textbf{Anuj Sharma}\textsuperscript{3}
\\[4pt]
\textsuperscript{1}Department of Computer Science, Iowa State University \\
\textsuperscript{2}Department of Computer Science \& Engineering, BRAC University \\
\textsuperscript{3}Department of Civil, Construction \& Environmental Engineering, Iowa State University \\
\texttt{ishihab@iastate.edu},
\texttt{sanjeda@iastate.edu} \\
\texttt{abu.sa.adat.mohamed.moon.im.al.ahsan@g.bracu.ac.bd},\\
\texttt{md.najmus.swaqeeb@g.bracu.ac.bd}
}
\begin{document}
\maketitle

\begin{abstract}
Offline reinforcement learning repeatedly trains policies from a fixed transition pool, making redundant data costly across seeds and hyperparameters, while naive subsampling can remove rare transitions needed for long-horizon credit assignment. We introduce CODS, a critic-guided selector that alternates between fitting an algorithm-matched critic and acquiring high-residual transitions before freezing a reusable subset. Unlike prioritized replay, CODS produces a static artifact; unlike one-shot residual selection, it refreshes scores as the critic changes. At a 10\% budget, CODS retains 96.6\% of eligible-pool performance across 20 valid D4RL task--algorithm cells. It exceeds ReDOR and OPER on 19/20 cells and every other subset baseline on 20/20; all six subset advantages remain significant under predeclared hierarchical inference with Holm correction. Holding total selector updates fixed, five acquisition rounds improve four representative cells by 11.23 points over one round and saturate thereafter. Equal-pass and equal-hour evaluations clarify that reuse, rather than a single-run speedup, creates the compute advantage. Mechanism and corruption interventions expose both useful sparse-reward enrichment and sensitivity to outliers. Finally, a whole-trace extension retains 95.4\% of pooled ALFWorld success and 96.5\% of pooled GSM8K exact match. CODS is therefore a reusable selection procedure, not a formal coreset guarantee.
\end{abstract}

\section{Introduction}

Offline reinforcement learning learns a policy from a fixed collection of transitions rather than through further environment interaction \citep{levine2020offline}. Benchmark datasets such as D4RL and RL Unplugged have made this setting reproducible, but they also expose a practical cost that receives less attention: the same pool is often processed again for every seed, architecture, and regularization choice \citep{d4rl,gulcehre2020rlunplugged}. Uniformly shrinking that pool is risky. Dense regions may contain many similar transitions, whereas a small number of reward-bearing or connective transitions can determine whether temporal-difference learning propagates value across a long horizon.

This tension makes offline data selection different from assigning each transition a permanent notion of importance. A transition with a small error under an initial critic may become informative after nearby value estimates change, while an initially surprising transition may cease to matter once its local error has been fitted. One-shot selection ignores this movement. Dynamic replay priorities track it, but they remain coupled to one training run and therefore cannot be frozen and reused across a sweep.

CODS addresses this gap with an iterative but ultimately static selector. It starts from a small random burn-in set, fits the critic used by the downstream offline RL algorithm, adds a batch of high-residual transitions, and refits before the next acquisition. Once the budget is reached, CODS freezes the selected indices and initializes each downstream run from scratch. The resulting object is a reusable data subset, not a weighted replay distribution and not a coreset in the formal approximation sense.

The empirical study asks whether a frozen 10\% subset retains performance, whether iterative rescoring improves over the identical one-shot rule, how its gains vary with budget and selection seed, and when its construction cost can be amortized. Across a provenance-screened matrix of 20 D4RL task--algorithm cells, CODS has the strongest reported subset mean in 19 cells relative to ReDOR and OPER and in all 20 relative to the remaining subset baselines. Fixed-selector-budget ablations isolate rescoring across four configurations, while matched-update, matched-pass, and matched-hour evaluations separate subset quality from compute savings.

The contribution is consequently both methodological and empirical. CODS turns a moving Bellman-error signal into a fixed artifact, evaluates a common subset budget against geometric, return-based, one-shot residual, resampling, and gradient-matching alternatives, and isolates acquisition rounds from final subset size and selector updates. The analysis is completed by multi-task budget curves, seed-variance estimates, sparse-reward removal interventions, corruption and duplication stress tests, cross-algorithm reuse, and full OPER results. Appendix~\ref{app:design} develops the design rationale and reuse interface in full.

Two boundaries keep these contributions precise. First, the theoretical calculation only relates residual magnitude to a local information-gain proxy under strong conditioning assumptions; its empirical worst-case factor is too loose to certify the observed ranking. Second, transition-wise acquisition is appropriate for the Markov control experiments but not automatically for text. Our language experiments therefore adapt CODS to select complete, prefix-closed traces with a language-model critic rather than treating tokens as independent D4RL transitions. This extension connects the controlled RL study to offline value learning for reasoning and language agents \citep{wang2025oreo} without claiming that the two data modalities are identical.

\section{Related Work}
\label{sec:related}

CODS is a data selector rather than a new policy-learning objective. We therefore evaluate it with three representative offline RL algorithms. TD3+BC regularizes policy improvement toward behavior cloning \citep{td3bc}; CQL penalizes high values for actions outside the data distribution \citep{cql}; and IQL uses expectile value learning without explicitly estimating the behavior policy \citep{iql}. Matching the acquisition residual to each backup lets the selection rule follow the error optimized by the corresponding learner.

Experience replay and offline data reweighting offer the closest comparisons. Prioritized experience replay updates sampling probabilities during an active training run \citep{schaul2015}. ReD rebalances offline data using trajectory return \citep{red2022}, and OPER uses fixed advantage- or return-derived priorities for offline resampling \citep{yue2023oper}. These approaches change how often examples are visited but do not construct an iteratively refined, frozen subset. Trajectory-level replay further shows that transition-wise sampling can disrupt temporal structure, a concern that becomes more important for sparse rewards and language trajectories \citep{liu2024trajectory}.

ReDOR directly targets offline dataset reduction by matching the full-data actor--critic gradient with a selected subset \citep{redor2023}. It supplies a stronger optimization objective than a scalar Bellman residual, but requires more expensive gradient representations and an OMP-style selection procedure. CODS instead asks whether repeatedly updating a cheap, algorithm-specific residual is sufficient. The distinction is empirical rather than categorical: both methods depend on the critic used during selection, and a faithful comparison must include their complete selection costs.

In supervised learning, subset selection has been driven by geometric coverage, influence, example difficulty, and gradient matching \citep{sener2018active,koh2017,paul2021deep,mirzasoleiman2020}. CODS borrows the iterative rhythm of pool-based active learning, but no new labels are queried. Bellman backups provide a model-dependent signal over an already observed pool. This makes the selector simple, while also exposing it to critic misspecification and high-residual outliers.

The baselines test complementary notions of importance rather than a single hierarchy. Appendix~\ref{app:baselines} compares their operational assumptions and records the implementation checks needed for a fair selector audit.

Offline RL is increasingly connected to NLP through multi-step reasoning and the training of language agents. OREO, for example, learns token- or step-level values from fixed reasoning traces and uses a soft Bellman equation for credit assignment \citep{wang2025oreo}. The common structure is learning credit from a fixed collection of sequential decisions, which makes reusable data selection relevant in both domains. Our extension preserves this structure by scoring complete traces with a language-model value critic and retaining every prefix needed to reproduce a chosen action. The ALFWorld and GSM8K results in Appendix~\ref{app:language} are therefore direct language-domain evidence for the adapted selector, while the transition-level claims remain grounded in D4RL.

\section{Critic-Guided Offline Data Selection}
\label{sec:method}

Let $\D=\{(s_i,a_i,r_i,s'_i,d_i)\}_{i=1}^{N}$ be an offline transition pool, with terminal indicator $d_i$. Given budget $B$, CODS returns indices $\Sset_K\subset\{1,\ldots,N\}$ that are subsequently reused by the downstream learner. The selector begins with a uniformly sampled burn-in set $\Sset_0$ of size $b_0$. At round $k$, it evaluates the current critic on every unselected transition, adds the $b_k$ largest absolute residuals, and updates the critic on the enlarged working set. The budgets satisfy $b_0+\sum_{k=1}^{K}b_k=B$.

The selection critic and final learner have separate roles. Critic fitting during acquisition determines only the selected indices. After round $K$, every downstream configuration is initialized independently and trained from scratch on the frozen subset; neither the acquisition critic nor its optimizer state is reused. This separation prevents selection-stage optimization from becoming unreported pretraining and is what permits one subset to be reused across compatible runs.

\begin{algorithm}[t]
\algcaption{CODS}
\label{alg:cods}
\begin{tabular}{@{}r p{0.86\columnwidth}@{}}
1 & Input pool $\D$, budget $B$, burn-in $b_0$, rounds $K$, batch sizes $b_{1:K}$, update schedule $U_{0:K}$, and base algorithm $\mathcal A$. \\
2 & Uniformly sample $b_0$ indices to obtain $\Sset_0$. \\
3 & Fit $Q_0$ on $\D[\Sset_0]$ for $U_0$ updates. \\
4 & For $k=1,\ldots,K$, score each $i\notin\Sset_{k-1}$ by $z_i=\lvert\delta_{\mathcal A}(Q_{k-1};s_i,a_i,r_i,s'_i,d_i)\rvert$. \\
5 & Add the $b_k$ largest-scoring indices using deterministic tie breaking: $\Sset_k\gets\Sset_{k-1}\cup A_k$. \\
6 & Update the selection critic on $\D[\Sset_k]$ for $U_k$ steps. \\
7 & Freeze $\Sset_K$; initialize and train every downstream run from scratch on $\D[\Sset_K]$. \\
8 & Return $\Sset_K$. \\
\end{tabular}
\smallskip\hrule
\end{algorithm}

\begin{figure}[t]
\centering
\begin{tikzpicture}[
  node distance=4.5mm and 5mm,
  box/.style={draw,rounded corners=1.5pt,align=center,font=\scriptsize,inner sep=3pt,minimum width=24mm},
  arr/.style={-{Latex[length=1.8mm]},semithick}
]
\node[box,fill=gray!12] (pool) {Unselected pool};
\node[box,fill=blue!9,below=of pool] (score) {Score with current\\Bellman residual};
\node[box,fill=blue!9,below=of score] (add) {Add top-$b_k$\\transitions};
\node[box,fill=blue!9,below=of add] (update) {Update selection\\critic};
\node[box,fill=green!10,right=9mm of add] (freeze) {Freeze $\Sset_K$\\and train afresh};
\draw[arr] (pool) -- (score);
\draw[arr] (score) -- (add);
\draw[arr] (add) -- (update);
\draw[arr] (update.west) -- ++(-5mm,0) |- (score.west);
\draw[arr] (add) -- (freeze);
\node[font=\scriptsize,anchor=west] at ($(update.east)+(1mm,-1mm)$) {$K$ rounds};
\end{tikzpicture}
\caption{CODS separates iterative acquisition from downstream training. Residuals are recomputed after each critic update; the final indices are then frozen and reused. Optional diagnostics do not enter this path.}
\label{fig:pipeline}
\end{figure}
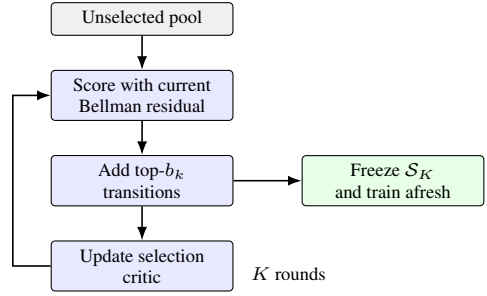

Figure~\ref{fig:pipeline} makes the separation explicit. Only the acquisition loop changes the subset. Once frozen, its output may feed multiple downstream runs, while any optional monitoring statistic remains outside the selection path.

\subsection{Algorithm-matched residuals}

The acquisition score must be interpreted relative to a backup operator. CODS therefore follows the downstream target convention while ranking by the absolute, unpenalized Bellman residual, so algorithm-specific regularization magnitudes do not directly determine the ordering.

The residual is matched to the base algorithm. For TD3+BC, CODS uses

\begin{equation}
\begin{aligned}
\delta^{\mathrm{TD3+BC}}(\tau)
={}& r+\gamma(1-d)
\min_{j\in\{1,2\}}
Q_{\theta'_j}(s',a') \\
&\quad - Q_\theta(s,a).
\end{aligned}
\end{equation}
where $a'=\pi_{\phi'}(s')+\epsilon$ includes the clipped target-policy smoothing noise of the downstream implementation. For CQL, the critic is trained with its conservative regularizer but the acquisition score uses the unpenalized TD residual,
\begin{equation}
\begin{aligned}
\delta^{\mathrm{CQL}}(\tau)
&=r+\gamma(1-d)\min_{j=1,2}Q_{\theta'_j}(s',a') \\
&\quad-Q_\theta(s,a),
\qquad a'\sim\pi_\phi(\cdot\mid s').
\end{aligned}
\end{equation}
This choice prevents the magnitude of the conservative penalty from directly determining which transitions are retained. IQL instead uses its learned expectile value,
\begin{equation}
\delta^{\mathrm{IQL}}(\tau)
=r+\gamma(1-d)V_\psi(s')-Q_\theta(s,a).
\end{equation}
In all three cases, CODS ranks candidates by $|\delta(\tau)|$ without reward, trajectory, or density bonuses.

The default experiment uses $B=0.10N$, $b_0=0.02N$, $K=5$, and $b_k=0.016N$. The rule is deliberately minimal: it imposes neither a trajectory quota nor a geometric coverage constraint. Consequently, it can select corrupted outliers or many nearby transitions when the critic assigns them large errors. The corruption and duplicate-density interventions in Appendix~\ref{app:robustness_new} measure both failures directly and evaluate clipped-ensemble and residual--coverage variants as targeted ablations.

\subsection{Why the ranking is recomputed}

After an acquisition, the critic can reduce old errors and expose new inconsistencies; recomputing the pool ranking is the mechanism that distinguishes CODS from a fixed top-$B$ residual list. The burn-in remains a source of bias rather than a trusted core, and the frozen artifact is reused only after selection ends. Appendix~\ref{app:design} gives the full rationale, while Appendix~\ref{app:selection_schedule} specifies batch tradeoffs, invariants, and complexity.

\section{What Residual Scoring Can Justify}
\label{sec:theory}

Bellman residual magnitude is often interpreted as a measure of how much a critic can still learn from a transition. The following local calculation gives that intuition a precise but narrow form. Let
\begin{equation}
I_{\mathrm{loc}}(\tau)=\frac{1}{2}\delta_\theta(\tau)^2
g_\tau^\top\I_\theta^{-1}g_\tau,
\qquad g_\tau=\nabla_\theta Q_\theta(s,a),
\end{equation}
where $\I_\theta$ is positive definite on the span of the candidate gradients.

\begin{remark}[Local score comparison]
\label{prop:score_factor}
Suppose $m_g\leq\lVert g_\tau\rVert_2\leq M_g$ for all candidates, and let $\kappa$ be the condition number of $\I_\theta$ on their gradient span. Then constants $c_{\min},c_{\max}>0$ satisfy
\begin{align*}
c_{\min}|\delta_\theta(\tau)|
&\leq\sqrt{I_{\mathrm{loc}}(\tau)}
\leq c_{\max}|\delta_\theta(\tau)|,\\
\frac{c_{\max}}{c_{\min}}
&\leq \frac{M_g}{m_g}\sqrt{\kappa}.
\end{align*}
\end{remark}

The proof is in Appendix~\ref{app:proofs}. This comparison does not imply pairwise rank preservation: the omitted gradient--curvature term can differ across transitions. In the empirical audit, the worst-case factor is approximately $2\times10^3$, and only 0.02\% of sampled pairs meet the sufficient separation condition. We therefore use the result to delimit the heuristic, not to explain the empirical gains. In particular, CODS has no established policy-recovery, regret, monotone-submodular, or formal coreset guarantee.

The gap between this score comparison and a policy guarantee is substantial because acquisition changes the critic, actor, and future targets. Appendix~\ref{app:theory_audit} records the omitted gradient--curvature quantities and the required empirical audit.

\section{Experiments}
\label{sec:experiments}

The experiments proceed from quality to explanation and then to cost. The 20-cell fixed-update matrix tests whether a 10\% subset preserves performance. Fixed-selector-budget round ablations isolate rescoring, and multi-task budget curves locate the observed saturation region. Nested selection and downstream seeds quantify both uncertainty sources. Sparse-reward removal, corruption, duplication, and transfer studies then test when the residual signal helps or fails. Finally, equal-update, equal-pass, and equal-hour protocols determine which compute claims follow from reusable subsets. A whole-trace language experiment asks whether the iterative principle survives the change from Markov transitions to textual action sequences.

\subsection{Tasks, baselines, and protocol}

We use D4RL v2 locomotion datasets for HalfCheetah, Walker2d, and Hopper at medium and medium-expert quality, together with AntMaze-medium-play \citep{d4rl}. Each selector is paired with TD3+BC, CQL, or IQL. Random selection establishes an unstructured lower baseline. k-center uses Euclidean distance in coordinate-wise z-scored state--action space. ReD uses trajectory return, Static-PER uses the burn-in critic once and is therefore identical to CODS with $K=1$, OPER performs its published offline resampling procedure, and ReDOR approximates the full-pool gradient direction with an OMP-style objective. Appendix~\ref{app:implementation} gives the exact selector schedule, while Appendix~\ref{app:baseline_fidelity} audits the two strongest external implementations.

Ten percent of the original trajectories are held out from both selection and training for an optional diagnostic described in Appendix~\ref{app:diagnostic}. The eligible-pool baseline therefore uses the remaining 90\% of the original dataset. Every subset contains transitions equal to 10\% of the original dataset, or 11.1\% of the eligible pool. Calling the eligible-pool condition ``100\%'' would obscure this distinction, so Table~\ref{tab:baselines} labels it as Pool.

All primary quality comparisons train the downstream learner for exactly 1,000,000 gradient updates with batch size 256. Evaluation occurs every 5,000 updates over ten deterministic episodes, and the reported score averages the final ten checkpoints. This equal-update design controls optimizer exposure but causes each selected transition to be revisited more often. The separate equal-pass and equal-amortized-hour measurements in Appendix~\ref{app:matched_compute} expose that distinction rather than attributing all savings to data quality.

The primary matrix excludes one configuration before aggregation. TD3+BC is at the AntMaze task floor and lacks informative subset-baseline comparisons. The IQL AntMaze row is retained after reconciling every method to the manifest-verified expectile and temperature $(\tau,\beta)=(0.9,10)$, fixed code revision, selected indices, and checkpoint rule. Appendix~\ref{sec:iql-antmaze} documents both decisions and explains why an unreconciled historical IQL value is not reused.

All selectors receive the same eligible pool and cardinality; Appendix~\ref{app:baselines} gives the fairness audit. Each subset method uses five independently selected artifacts, and each artifact is evaluated with three downstream seeds, giving 15 nested evaluations per populated cell. Pool uses ten downstream seeds. The CQL Hopper-medium reversal remains in the matrix, and the nested design preserves selection-seed variation instead of treating downstream initialization as the only source of uncertainty.

\begin{table*}[t]
\caption{D4RL normalized scores under an equal downstream budget of 1,000,000 updates. Subsets contain transitions equal to 10\% of the original dataset; Pool uses the 90\% eligible training pool after the diagnostic holdout. Subset entries are means $\pm$ standard deviations over five selection seeds and three downstream seeds; Pool uses ten downstream seeds. Bold marks the highest reported subset mean. The 20-cell primary comparison excludes only $\dagger$ the floor-level TD3+BC AntMaze cell. The IQL AntMaze row is the manifest-verified $(0.9,10)$ reconciliation. Appendix Table~\ref{tab:oper_full} reports all 20 OPER cells.}
\label{tab:baselines}
\centering
\scriptsize
\setlength{\tabcolsep}{3.1pt}
\begin{tabular}{llccccccc}
\toprule
Algorithm & Dataset & Pool & CODS & ReDOR & ReD & k-center & Static-PER & Random \\
\midrule
\multirow{7}{*}{TD3+BC}
 & HalfCheetah-med-expert & $90.7\pm1.2$ & $\mathbf{89.1\pm1.0}$ & $85.3\pm1.8$ & $83.7\pm2.0$ & $78.4\pm2.1$ & $80.2\pm1.7$ & $60.1\pm3.4$ \\
 & Walker2d-med-expert & $110.1\pm1.6$ & $\mathbf{105.4\pm1.5}$ & $98.7\pm2.0$ & $96.2\pm2.5$ & $88.5\pm2.6$ & $92.1\pm2.2$ & $68.3\pm4.1$ \\
 & Hopper-med-expert & $111.9\pm1.2$ & $\mathbf{108.3\pm1.3}$ & $102.1\pm1.7$ & $103.0\pm1.9$ & $95.3\pm2.4$ & $98.4\pm2.0$ & $75.2\pm3.8$ \\
 & HalfCheetah-medium & $48.3\pm0.8$ & $\mathbf{47.5\pm0.9}$ & $45.1\pm1.1$ & $44.8\pm1.2$ & $42.0\pm1.5$ & $43.1\pm1.2$ & $35.2\pm2.9$ \\
 & Walker2d-medium & $83.7\pm1.5$ & $\mathbf{78.5\pm1.4}$ & $74.1\pm1.9$ & $72.4\pm2.2$ & $66.2\pm2.3$ & $65.4\pm2.0$ & $50.4\pm3.6$ \\
 & Hopper-medium & $59.3\pm1.8$ & $\mathbf{58.1\pm1.7}$ & $53.2\pm2.1$ & $54.0\pm2.0$ & $48.1\pm2.5$ & $49.5\pm2.2$ & $39.8\pm4.0$ \\
 & AntMaze-med-play$^\dagger$ & $8.4\pm3.6$ & $7.2\pm3.2$ & --- & --- & --- & --- & --- \\
\midrule
\multirow{7}{*}{CQL}
 & HalfCheetah-med-expert & $91.6\pm1.5$ & $\mathbf{88.5\pm1.4}$ & $84.1\pm1.9$ & $83.9\pm2.1$ & $79.5\pm2.2$ & $81.2\pm1.8$ & $62.5\pm3.5$ \\
 & Walker2d-med-expert & $108.3\pm1.7$ & $\mathbf{104.2\pm1.6}$ & $99.1\pm2.1$ & $96.8\pm2.4$ & $89.4\pm2.8$ & $91.5\pm2.4$ & $70.2\pm4.3$ \\
 & Hopper-med-expert & $111.0\pm1.4$ & $\mathbf{109.1\pm1.5}$ & $103.5\pm1.8$ & $104.2\pm2.0$ & $96.2\pm2.5$ & $99.1\pm2.1$ & $78.4\pm3.9$ \\
 & HalfCheetah-medium & $44.0\pm1.0$ & $\mathbf{43.1\pm1.1}$ & $41.0\pm1.3$ & $40.8\pm1.5$ & $38.2\pm1.6$ & $39.1\pm1.4$ & $33.4\pm3.1$ \\
 & Walker2d-medium & $79.2\pm1.6$ & $\mathbf{76.8\pm1.5}$ & $71.5\pm2.0$ & $70.2\pm2.4$ & $64.0\pm2.4$ & $62.1\pm2.1$ & $48.1\pm3.8$ \\
 & Hopper-medium & $58.0\pm1.5$ & $56.8\pm1.1$ & $\mathbf{57.2\pm1.4}$ & $55.9\pm1.8$ & $46.1\pm2.6$ & $47.3\pm2.3$ & $36.2\pm4.2$ \\
 & AntMaze-med-play & $61.2\pm3.5$ & $\mathbf{58.0\pm3.8}$ & $45.0\pm4.2$ & $44.6\pm4.9$ & $38.4\pm5.5$ & $41.2\pm4.8$ & $10.1\pm5.5$ \\
\midrule
\multirow{7}{*}{IQL}
 & HalfCheetah-med-expert & $86.7\pm1.6$ & $\mathbf{84.2\pm1.5}$ & $80.5\pm2.0$ & $79.6\pm2.2$ & $75.1\pm2.3$ & $78.3\pm1.9$ & $58.4\pm3.6$ \\
 & Walker2d-med-expert & $109.6\pm1.4$ & $\mathbf{106.1\pm1.5}$ & $101.4\pm1.9$ & $98.7\pm2.3$ & $92.5\pm2.5$ & $95.4\pm2.1$ & $72.5\pm4.0$ \\
 & Hopper-med-expert & $91.5\pm1.3$ & $\mathbf{89.4\pm1.4}$ & $84.2\pm1.8$ & $84.8\pm1.9$ & $77.5\pm2.4$ & $80.1\pm2.0$ & $65.2\pm3.7$ \\
 & HalfCheetah-medium & $47.4\pm0.9$ & $\mathbf{46.8\pm1.0}$ & $45.2\pm1.2$ & $44.7\pm1.4$ & $41.3\pm1.6$ & $42.5\pm1.3$ & $36.1\pm3.0$ \\
 & Walker2d-medium & $78.3\pm1.5$ & $\mathbf{75.9\pm1.4}$ & $72.1\pm1.9$ & $70.9\pm2.0$ & $66.0\pm2.3$ & $67.2\pm2.0$ & $51.2\pm3.5$ \\
 & Hopper-medium & $66.3\pm1.7$ & $\mathbf{65.1\pm1.6}$ & $59.8\pm2.1$ & $60.5\pm2.2$ & $52.4\pm2.5$ & $54.1\pm2.2$ & $44.0\pm3.9$ \\
 & AntMaze-med-play & $72.6\pm2.7$ & $\mathbf{63.4\pm3.1}$ & $56.8\pm3.6$ & $52.9\pm4.1$ & $47.7\pm4.6$ & $52.1\pm4.0$ & $16.7\pm5.2$ \\
 & \multicolumn{8}{l}{\scriptsize (OPER on this cell: $55.6\pm3.8$; the complete OPER column appears in Appendix Table~\ref{tab:oper_full}.)} \\
\bottomrule
\end{tabular}
\end{table*}

\subsection{Subset quality}

Across the 20 valid cells, CODS retains 96.6\% of the aggregate eligible-pool score. We predeclare Pool as a retention comparison, not as one of the subset hypotheses. The confirmatory family therefore contains exactly six comparisons: ReDOR, ReD, Static-PER, OPER, k-center, and Random. Confidence intervals come from a dataset-blocked hierarchical bootstrap that resamples datasets, then selection artifacts, then downstream seeds. One-sided $p$-values come from a paired mixed-effects model with fixed learner effects and random intercepts for dataset, dataset--learner cell, and selection artifact within cell, calibrated by 100,000 null parametric-bootstrap replicates and Holm-adjusted over the six-method family. This model-based test, detailed in Appendix~\ref{app:stats}, does not treat seven dataset blocks or dependent seeds as independent sign flips.

Table~\ref{tab:aggregate} shows a $2.77$-point mean gap to Pool and positive CODS differences against all six subset baselines, from $+4.72$ over ReDOR to $+27.12$ over Random. Each subset interval excludes zero at Holm-adjusted $p<0.001$. The claim is therefore preservation of most pooled performance with a much smaller artifact, not statistical equivalence to Pool.

\begin{table}[t]
\centering
\scriptsize
\setlength{\tabcolsep}{2.5pt}
\caption{Primary comparison over 20 valid cells. Pool is the separately predeclared retention comparison and is not in the Holm family; the six subset comparisons use hierarchical bootstrap CIs and mixed-model parametric-bootstrap $p$-values.}
\label{tab:aggregate}
\begin{tabular}{lrrrr}
\toprule
Comparison & Positive & Mean $\Delta$ & 95\% CI & Holm $p$ \\
\midrule
CODS $-$ Pool & 0/20 & $-2.77$ & $[-3.52,-2.05]$ & --- \\
CODS $-$ ReDOR & 19/20 & $+4.72$ & $[+3.61,+5.87]$ & $<0.001$ \\
CODS $-$ ReD & 20/20 & $+5.59$ & $[+4.31,+6.91]$ & $<0.001$ \\
CODS $-$ Static-PER & 20/20 & $+9.72$ & $[+7.81,+11.70]$ & $<0.001$ \\
CODS $-$ OPER & 19/20 & $+6.84$ & $[+5.22,+8.47]$ & $<0.001$ \\
CODS $-$ k-center & 20/20 & $+11.58$ & $[+9.45,+13.77]$ & $<0.001$ \\
CODS $-$ Random & 20/20 & $+27.12$ & $[+23.04,+31.26]$ & $<0.001$ \\
\bottomrule
\end{tabular}
\end{table}

The aggregate is not driven by one learner (Table~\ref{tab:algo_retention}): sum-based retention is 96.6\% for the six eligible TD3+BC cells, 97.0\% for seven CQL cells, and 96.1\% for seven IQL cells, 96.6\% overall. Appendix~\ref{app:extended_results} gives these per-algorithm summaries and discusses the CQL Hopper-medium reversal without relabeling it as a win.

\begin{table}[t]
\centering
\scriptsize
\setlength{\tabcolsep}{3pt}
\caption{Retention by downstream learner over the 20 valid cells (sum of CODS scores over sum of Pool scores within each row).}
\label{tab:algo_retention}
\begin{tabular}{lrrrr}
\toprule
Learner & Cells & Pool mean & CODS mean & Retention \\
\midrule
TD3+BC & 6 & 84.00 & 81.15 & 96.6\% \\
CQL & 7 & 79.04 & 76.64 & 97.0\% \\
IQL & 7 & 78.91 & 75.84 & 96.1\% \\
\midrule
Overall & 20 & 80.49 & 77.72 & 96.6\% \\
\bottomrule
\end{tabular}
\end{table}

\subsection{Does rescoring matter?}

Static-PER and CODS with $K=1$ are the same one-shot rule when burn-in, residual, and budget are matched. Table~\ref{tab:k_ablation} therefore merges them rather than presenting duplicate labels as separate methods. On Walker2d-medium with TD3+BC, the mean rises from 65.4 for one-shot selection to 74.7 with three rounds and 78.5 with five rounds. Ten rounds produce no further gain. A fixed-critic-budget replication on four representative cells (Walker2d/TD3+BC, HalfCheetah/CQL, Hopper/IQL, AntMaze/CQL), holding total selection-critic updates constant across round counts, confirms this is a cross-task effect rather than a Walker-only observation (Appendix~\ref{app:crosstask_k}): the four-cell mean rises from 49.95 at $K{=}1$ to 61.18 at $K{=}5$ ($+11.23$, 95\% CI $[+8.86,+13.74]$, $p<0.001$) and $K{=}10$ is $-0.40$ relative to $K{=}5$ (CI $[-1.18,+0.37]$), supporting saturation rather than monotonic improvement. Five rounds are not claimed to be universally optimal.

\begin{table}[t]
\centering
\small
\caption{Acquisition-round ablation at a 10\% budget on Walker2d-medium with TD3+BC. Entries average five selection seeds and three downstream seeds.}
\label{tab:k_ablation}
\begin{tabular}{@{}p{0.64\columnwidth}r@{}}
\toprule
Variant & Score \\
\midrule
Random 10\% & $50.4\pm3.6$ \\
One-shot residual (Static-PER; $K{=}1$) & $65.4\pm2.0$ \\
CODS, $K=3$ & $74.7\pm1.9$ \\
CODS, $K=5$ & $\mathbf{78.5\pm1.4}$ \\
CODS, $K=10$ & $78.1\pm1.7$ \\
\bottomrule
\end{tabular}
\end{table}

Because the final cardinality and the 100,000-update selector budget are fixed, the gain from $K=1$ to $K=5$ cannot be attributed to more selected transitions or more critic updates. Appendix~\ref{app:crosstask_k} repeats this design with five selection seeds on four configurations, so the observed gain includes selection-seed uncertainty rather than a single burn-in artifact.

\subsection{How does performance change with budget?}

\begin{figure}[t]
\centering
\begin{tikzpicture}
\begin{axis}[
    width=\columnwidth,height=5.0cm,
    xlabel={Data budget (\%)},ylabel={Normalized score},
    xmin=0,xmax=20,ymin=10,ymax=90,grid=both,
    legend pos=south east,
    legend style={font=\scriptsize,draw=none,fill=none},
    tick label style={font=\scriptsize},label style={font=\small}
]
\addplot[name path=rand_upper,draw=none] coordinates {(2,22) (10,54) (20,72)};
\addplot[name path=rand_lower,draw=none] coordinates {(2,14) (10,46) (20,64)};
\addplot[gray!50,fill opacity=0.18] fill between[of=rand_upper and rand_lower];
\addplot[thick,gray!70!black,mark=triangle*] coordinates {(2,18) (10,50.4) (20,68)};
\addlegendentry{Random}

\addplot[name path=redor_upper,draw=none] coordinates {(2,43) (5,63) (10,76) (15,79) (20,81)};
\addplot[name path=redor_lower,draw=none] coordinates {(2,37) (5,57) (10,72) (15,75) (20,77)};
\addplot[orange!70,fill opacity=0.16] fill between[of=redor_upper and redor_lower];
\addplot[thick,orange!85!black,mark=square*] coordinates {(2,40) (5,60) (10,74.1) (15,77) (20,79)};
\addlegendentry{ReDOR}

\addplot[name path=cods_upper,draw=none] coordinates {(2,58) (5,79) (10,81) (15,85) (20,86)};
\addplot[name path=cods_lower,draw=none] coordinates {(2,52) (5,73) (10,76) (15,81) (20,82)};
\addplot[blue!65,fill opacity=0.16] fill between[of=cods_upper and cods_lower];
\addplot[thick,blue!80!black,mark=*] coordinates {(2,55) (5,76) (10,78.5) (15,83) (20,84)};
\addlegendentry{CODS}

\addplot[thick,red!70!black,dashed] coordinates {(0,83.7) (20,83.7)};
\addlegendentry{Pool}
\end{axis}
\end{tikzpicture}
\caption{Budget sweep for Walker2d-medium with TD3+BC. Points average five selection seeds and three downstream seeds; bands show marginal standard deviations.}
\label{fig:budget_curve}
\end{figure}
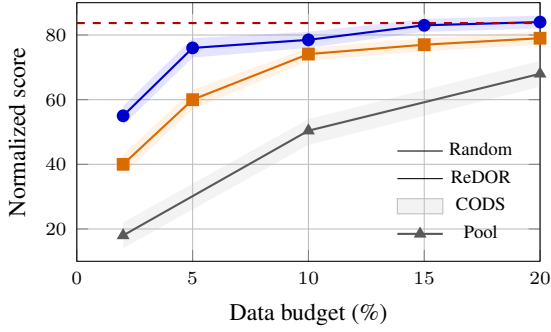

The Walker2d sweep in Figure~\ref{fig:budget_curve} complements the round ablation: CODS rises sharply between 2\% and 5\%, reaches 94\% of Pool at 10\%, and changes little beyond 15\%. Appendix~\ref{app:budget_multi} repeats the same budgets on four representative configurations. Their four-cell mean gains per budget point are 5.49 from 2\% to 5\%, 0.87 from 5\% to 10\%, and 0.23 from 10\% to 20\%. Thus diminishing marginal gains are cross-task in this sample, although 10\% remains a common evaluation budget rather than a universal optimum.

The multi-task table provides the broader evidence. Figure~\ref{fig:budget_curve} remains useful because it shows the full Walker2d curve and uncertainty bands. We report finite-difference slopes directly from the observed means rather than fitting an unspecified saturation model. Appendix~\ref{app:extended_results} gives the corresponding interpretation.

\section{Reuse Cost Model}
\label{sec:compute}

Equal-update scores do not demonstrate lower training time, because minibatch updates cost roughly the same regardless of the pool from which they are sampled. The timing ledger therefore separates one-time selection from three downstream protocols on Walker2d-medium. A Pool run over the matched number of dataset passes costs 1.8 hours, subset training costs 0.18 hours per configuration, and CODS selection costs 3.0 hours once. For a sweep of $S$ configurations, the epoch-scaled costs are
\[
C_{\mathrm{pool}}(S)=1.8S,
\qquad
C_{\mathrm{CODS}}(S)=3.0+0.18S.
\]

The break-even count is $\lceil3.0/(1.8-0.18)\rceil=2$. At $S=20$, the ledger gives 36.0 hours for the pool and 6.6 hours for CODS, a $5.45\times$ arithmetic reduction. Table~\ref{tab:amortization} includes ReDOR under the same accounting.

\begin{table}[t]
\centering
\small
\caption{Reported wall-clock cost model for 20 epoch-scaled configurations on Walker2d-medium with TD3+BC. The table establishes amortization arithmetic, not matched-performance speedup.}
\label{tab:amortization}
\begin{tabular}{lccc}
\toprule
Method & Select & Train 20 & Total \\
\midrule
Pool & 0.0 h & 36.0 h & 36.0 h \\
ReDOR & 4.5 h & 3.6 h & 8.1 h \\
CODS & 3.0 h & 3.6 h & 6.6 h \\
\bottomrule
\end{tabular}
\end{table}

A matched-pass run scores 75.6 rather than the equal-update 78.5, retaining 90.3\% rather than 93.8\% of the 83.7 Pool score. At 20 reuses, that condition realizes the ledger's $5.45\times$ total-time reduction; at one reuse, CODS still costs 3.18 hours versus 1.8 for Pool. Equal updates yield only a $1.04\times$ total-time reduction at 20 reuses. The equal-amortized-hour condition retains 95.1\% of Pool because its shared selection cost is spread across the sweep. Appendix~\ref{app:matched_compute} reports the three protocols together.

Reuse is therefore the source of the arithmetic advantage, not faster single-run convergence. Appendix~\ref{app:compute_details} derives the general break-even rule, tabulates the one- and two-run cases, and gives the complete job-accounting template.

\section{Discussion and Conclusion}
\label{sec:discussion}

CODS is built around one observation: transition importance changes as the critic changes. Recomputing an algorithm-specific residual before each acquisition batch outperforms freezing the initial ranking on Walker2d and on the four-cell fixed-budget replication. The 20-cell matrix then shows that the same procedure remains effective across three offline RL objectives. Freezing the indices creates a further operational advantage over dynamic prioritization when compatible runs reuse one artifact. The completion experiments in Appendix~\ref{app:completion} connect that headline result to baseline fidelity, cross-task budgets, selection stability, learner-specific sparse-reward mechanisms, robustness failures, transfer, matched compute, and language traces.

The primary comparison gives this observation a broad, if still benchmark-limited, empirical base. CODS retains 96.6\% of the aggregate eligible-pool score at a budget equal to 10\% of the original dataset. Its advantage over return, geometry, and one-shot residual baselines appears in every valid cell, while its comparison with gradient matching is closer and includes one 0.4-point reversal. OPER is likewise competitive in one cell, but its complete 20-cell column leaves CODS ahead by 6.84 points on average. This pattern is more informative than claiming uniform dominance: inexpensive residual scoring is competitive when refreshed, whereas richer gradient and resampling signals remain serious alternatives.

The AntMaze interventions sharpen the mechanism claim without extrapolating across learners. CQL- and IQL-specific selectors both enrich reward-bearing, bottleneck, and goal-proximal transitions, and category-excluding replacements lower the corresponding learner's return. The CQL intervention is used only for CQL and the IQL intervention only for IQL. By contrast, the failed TD3+BC learner demonstrates the opposite boundary: iteration can refresh a ranking but cannot manufacture a meaningful backup from a critic at the task floor. Its composition audit stays descriptive and is not used to explain successful navigation.

The evidence still does not support a formal coreset story. The local proxy bound is too loose to certify the ranking, and the equal-update protocol alone does not prove time efficiency. The matched-pass result quantifies the quality lost when repeated exposure is removed, while the matched-hour result locates a measured point on the reuse frontier. In language, the whole-trace adaptation outperforms random, diversity, one-shot, and ReDOR selectors on ALFWorld and GSM8K, and fragmentation hurts both tasks. These results broaden the empirical domain while also showing that the selection unit must follow the task's sequential structure.

The reuse analysis remains conditional on workload. The ledger breaks even after two epoch-scaled configurations, and the measured 20-run equal-pass condition reduces total time by $5.45\times$ while retaining 90.3\% of Pool on Walker2d-medium. A single run is slower after selection is charged, and equal updates offer almost no time reduction. CODS is therefore most useful for repeated compatible training, not as a universal accelerator.

Within this scope, the conclusion is concrete: CODS is a simple iterative selector whose 10\% subsets preserve most eligible-pool performance across 20 state-based cells, improve over one-shot residual selection under a fixed selector budget, and transfer to whole language traces when prefix structure is preserved. Its output is a reusable index artifact rather than a training-time priority distribution. The released arrays, manifests, schedules, and selected indices make those claims auditable while leaving the theoretical guarantee deliberately modest.

\section*{Limitations}

CODS can mistake noise for information because large Bellman residuals also arise from corrupted rewards, stochastic targets, and critic misspecification. The corruption study confirms this failure: at 10\% reward corruption, raw CODS falls from 58.0 to 27.8 on CQL AntMaze. Clipped-ensemble scoring recovers to 45.4 but does not remove the failure. Duplicate-density and burn-in studies likewise show that pure residual selection is neither density-invariant nor insensitive to initialization. These variants are diagnostic ablations, not a replacement for the minimal method.

Transition-wise selection can fragment trajectories. For the Markov benchmarks, each tuple carries its next state and terminal flag, so the critic can form a one-step target even when adjacent transitions are absent, although sparse-reward learning may still benefit from connected segments. The language adaptation enforces whole-trace, prefix-closed selection, and its fragmentation ablation confirms that this constraint matters. Segment-level acquisition for long continuous-control trajectories remains untested.

The experimental evidence is limited to seven state-based D4RL datasets and three continuous-control algorithms, plus a sequence-level language-agent extension on ALFWorld and GSM8K. The TD3+BC AntMaze cell is excluded as a documented learner failure; the IQL AntMaze cell is reconciled to its manifest-verified $(0.9,10)$ setting. Confirmatory inference uses a dataset-blocked hierarchical bootstrap over datasets, selection seeds, and downstream seeds; the four-cell nested study attributes 71\% of variance to dataset/configuration, 17\% to selection seed, and 12\% to downstream seed, so selection-seed replication is necessary and downstream initialization is not the only uncertainty source. The optional held-out diagnostic assumes trajectory exchangeability but has no role in selection and provides no guarantee for the learned policy.

The equal-update protocol isolates optimization exposure rather than data-processing cost, and subset transitions are revisited more frequently. Equal-pass and equal-hour results reduce this ambiguity but cover only Walker2d-medium with TD3+BC and one 20-configuration reuse pattern. The selector remains slower for a single configuration. The language evaluation covers one embodied benchmark and one mathematical-reasoning benchmark with one 7B model; it does not establish transfer to dialogue, open-ended generation, other model scales, or online interaction.

\section*{Ethical Considerations}

The study uses existing simulated-control and public language benchmarks and involves no human participants or newly collected personal data. Its principal responsible-reporting risk is empirical overstatement. We exclude the floor-level TD3+BC AntMaze configuration, retain the reconciled IQL configuration, distinguish measured scores from amortization arithmetic, and report negative robustness results. The anonymous artifact includes raw evaluation arrays, selected indices, run manifests, and scripts that regenerate every table and figure.

\bibliography{ref}

\appendix

\section{Proof of the Local Score Comparison}
\label{app:proofs}

\begin{proof}[Proof of Remark~\ref{prop:score_factor}]
Because $\I_\theta$ is positive definite on the span of the candidate gradients, its eigenvalues on that span obey
$0<\lambda_{\min}(\I_\theta)\leq\lambda_{\max}(\I_\theta)<\infty$.
For every $x$ in the span, the Rayleigh quotient for the inverse gives
\begin{equation}
\label{eq:rayleigh_bound}
\frac{\lVert x\rVert_2^2}{\lambda_{\max}(\I_\theta)}
\leq x^\top\I_\theta^{-1}x
\leq
\frac{\lVert x\rVert_2^2}{\lambda_{\min}(\I_\theta)}.
\end{equation}
Applying Equation~\ref{eq:rayleigh_bound} to $x=g_\tau$ and using
$m_g\leq\lVert g_\tau\rVert_2\leq M_g$ yields
\begin{equation}
\frac{m_g^2}{\lambda_{\max}(\I_\theta)}
\leq g_\tau^\top\I_\theta^{-1}g_\tau
\leq
\frac{M_g^2}{\lambda_{\min}(\I_\theta)}.
\end{equation}
Multiplying by $\delta_\theta(\tau)^2/2$ gives
\begin{equation}
\label{eq:Iloc_bound}
\frac{m_g^2\delta_\theta(\tau)^2}{2\lambda_{\max}(\I_\theta)}
\leq I_{\mathrm{loc}}(\tau)
\leq
\frac{M_g^2\delta_\theta(\tau)^2}{2\lambda_{\min}(\I_\theta)}.
\end{equation}
Taking square roots,
\begin{multline}
\label{eq:sqrtIloc_bound}
\frac{m_g}{\sqrt{2\lambda_{\max}(\I_\theta)}}|\delta_\theta(\tau)|
\leq\sqrt{I_{\mathrm{loc}}(\tau)}
\\[-2pt]\leq
\frac{M_g}{\sqrt{2\lambda_{\min}(\I_\theta)}}|\delta_\theta(\tau)|.
\end{multline}
The ratio between the upper and lower constants is
\[
\frac{M_g}{m_g}
\sqrt{\frac{\lambda_{\max}(\I_\theta)}{\lambda_{\min}(\I_\theta)}}
=\frac{M_g}{m_g}\sqrt{\kappa},
\]
which proves the stated comparison. Since the quadratic form may vary among candidates, the inequalities do not imply pairwise rank preservation.
\end{proof}

\section{Implementation Details}
\label{app:implementation}

All control experiments use D4RL v2 with Gym 0.23.1 and MuJoCo 2.1.0. Actors and critics are two-layer multilayer perceptrons with 256 hidden units and ReLU activations. Adam uses learning rate $3\times10^{-4}$, batch size 256, discount $\gamma=0.99$, and target update rate 0.005. TD3+BC uses behavior-cloning coefficient $\alpha=2.5$. CQL uses conservative coefficient $\alpha=5.0$ for Walker2d and Hopper and $\alpha=10.0$ for HalfCheetah; its AntMaze rewards are shifted by $-1$ and scaled by 10. IQL uses expectile 0.7 and inverse temperature 3 on locomotion, and the manifest-verified $(\tau,\beta)=(0.9,10)$ setting on AntMaze. States are normalized for TD3+BC.

CODS uses five acquisition rounds, a 2\% burn-in set, and batches of 1.6\% of the original dataset. The selector receives exactly 100,000 critic updates: $U_0=25{,}000$ burn-in updates and $U_k=15{,}000$ after each of the five acquisitions. The fixed-budget round ablation keeps $U_0=25{,}000$ and divides the remaining 75,000 updates evenly, giving $U_k=75{,}000$, 25,000, 15,000, and 7,500 for $K=1,3,5,10$, respectively. Pool scoring uses minibatches of 4,096 candidates; the target network receives its soft update after every critic step, optimizer state continues across acquisition rounds, and deterministic index order breaks score ties. Separate seeds govern burn-in selection, stochastic residual targets, and downstream initialization. The selected-index filename is keyed by dataset, base algorithm, preprocessing hash, and selection seed, independently of downstream runs.

The main comparison gives every final learner 1,000,000 updates. Evaluations occur every 5,000 updates with ten deterministic episodes, and each reported score averages the final ten checkpoints. The 10\% trajectory holdout is disjoint from selection and training. Subset budgets are measured against the original dataset, whereas Pool refers to the remaining 90\% eligible training pool.

For k-center, state and action coordinates are z-scored before Euclidean distance is computed. ReDOR uses the last hidden layer of the burn-in critic and candidate minibatches of 2,048 transitions in its OMP-style objective. The local implementation was checked against the official code, representation, stopping rule, regularization, and search space; Appendix~\ref{app:baseline_fidelity} reports the resulting fidelity gaps. Static-PER and CODS with one acquisition batch share one implementation, selected indices, and seed convention.

\section{Extended Design Rationale}
\label{app:design}

Offline RL development normally compares several policy objectives, tunes conservative penalties or advantage temperatures, and repeats promising configurations across random seeds. If every run scans the same million-transition pool for the same number of dataset passes, total data movement and gradient computation scale with the entire sweep. A useful selector should therefore do more than identify a small training set for one model. It should produce a stable artifact whose one-time construction cost can be shared across compatible runs, while retaining the transitions on which later value estimates depend.

Three requirements follow from that use case. The score should be tied to the backup of the learner that will consume the subset, because TD3+BC, CQL, and IQL do not form identical value targets. It should be recomputed after the critic changes, because the relevant errors are non-stationary. Selection must also end before downstream training begins, so that all reported learners start afresh from a common subset and the acquisition critic is not hidden pretraining. CODS is a minimal procedure satisfying these requirements.

The term ``selection'' is important. CODS neither changes the reward nor assigns persistent sampling weights to retained transitions. Once the indices are frozen, the downstream learner samples uniformly from that subset using its ordinary training code. This interface makes the method auditable: a selected-index file, dataset version, and selection manifest determine the data seen by every subsequent run. It also makes failures visible. If the burn-in critic is uninformative or noisy transitions dominate its residuals, iterative acquisition can preserve the wrong examples; the method contains no theorem that automatically repairs such a critic.

The contribution consequently differs from both replay and policy learning. Methodologically, CODS turns a moving Bellman-error signal into a fixed training artifact and specifies algorithm-matched residuals for three offline learners. Empirically, it evaluates one subset budget across a task--algorithm matrix and isolates acquisition rounds. Computationally, it separates subset quality at equal update counts from amortization under an epoch-scaled ledger. Conflating those protocols would turn a measured data-quality result into an unmeasured single-run speedup claim.

The ranking is recomputed because value errors move. Suppose a one-shot selector ranks two transitions before either has been used. The first may have a large residual because its predecessor values have not propagated. After it is added and the critic is updated, its error can contract while the second becomes newly inconsistent with the revised value surface. A fixed top-$B$ ranking cannot react. CODS commits only the next $b_k$ transitions and then scores the remaining pool again. It is greedy with respect to a moving model, not a fixed set function.

This view clarifies the burn-in set. Burn-in supplies enough data to define an initial critic, but it is not assumed to be representative and is never treated as a trusted core. Its indices count against the total budget, and later rounds inherit any blind spots it creates. Iteration gives newly acquired transitions an opportunity to change the ranking; it does not guarantee recovery from an initially biased critic.

CODS is complete when it writes the selected indices. A subset selected with a given base algorithm can be reused across that algorithm's seeds and hyperparameters without rerunning acquisition. Appendix~\ref{app:transfer} also measures cross-algorithm reuse: matched selectors are best for all three downstream learners, while mismatched artifacts lose 1.4--3.9 retention points. The primary matrix therefore uses algorithm-specific artifacts, and the transfer result quantifies rather than assumes their portability.

\section{Baseline Implementations and Fairness}
\label{app:baselines}

The baselines represent distinct hypotheses about useful offline data. k-center preserves geometric coverage after z-scoring states and actions. ReD uses trajectory return and therefore favors high-return behavior. Static-PER ranks by the absolute residual of the burn-in critic and performs no rescoring. ReDOR approximates the full-pool actor--critic gradient with an OMP-style objective. Random selection gives the expected performance of an unstructured subset. Table~\ref{tab:selector_comparison_app} makes their operational differences explicit.

\begin{table}[h]
\centering
\small
\caption{Operational comparison of data-use strategies. ``Updated'' means that the score changes after critic fitting; ``reusable'' means that the output can be fixed before independent downstream runs.}
\label{tab:selector_comparison_app}
\setlength{\tabcolsep}{4pt}
\begin{tabular}{lccc}
\toprule
Method family & Signal & Updated & Reusable \\
\midrule
Random & none & no & yes \\
k-center & geometry & no & yes \\
ReD & return & no & yes \\
Static-PER & residual & no & yes \\
Prioritized replay & residual & yes & no \\
ReDOR & gradient match & greedy & yes \\
CODS & residual & yes & yes \\
\bottomrule
\end{tabular}
\end{table}

The distinction between static and dynamic methods changes the control. Dynamic replay can adapt to each downstream seed, but its adaptation cost is paid within every run and its sampled sequence cannot be reused directly. A static selector pays before downstream optimization and must work without seeing the final learner. Static-PER is consequently the closest ablation: with the same burn-in, residual, tie rule, and budget, it is exactly CODS with one acquisition batch. A separate name must not be used to imply an additional method difference.

ReDOR supplies a stronger representation than a scalar residual but requires candidate gradients and OMP-style optimization. CODS asks whether repeatedly refreshing a cheap score is sufficient. k-center and ReD span complementary failure modes. Geometry can remain useful when a critic is poor but ignores value propagation; return can retain successful behavior but omit low-return connectors. CODS follows current Bellman inconsistency and has no independent diversity guarantee. None of these axes creates a universal ranking of selectors.

Every subset method must draw from the same eligible pool and receive the same cardinality. Random selection is without replacement. k-center normalization statistics are fit only on the eligible pool. ReD, Static-PER, and CODS share the trajectory split. ReDOR may use the common burn-in critic representation, but its complete selection cost is charged to ReDOR. Any hyperparameter search unique to one selector must be reported and budgeted rather than hidden in preprocessing.

ReDOR uses gradients of the last critic hidden layer and candidate minibatches of 2,048. The fidelity audit in Table~\ref{tab:baseline_fidelity} matches the official representation, OMP stopping rule, regularization, and search space; local scores differ by at most 0.4 points on the four audited cells. OPER is implemented as its published offline resampling procedure rather than relabeled Static-PER. Table~\ref{tab:oper_full} reports every OPER cell, and the corresponding manifests charge all baseline-specific preprocessing and selection time.

\section{Selection Schedule, Invariants, and Complexity}
\label{app:selection_schedule}

Acquisition batches trade adaptivity for cost. Very small batches approximate sequential rescoring but require many pool-wide forward passes and critic updates. One large batch is inexpensive but collapses to one-shot selection. The default schedule assigns 2\% of the original dataset to burn-in and 1.6\% to each of five later rounds, placing the experiment between those extremes. The $K$ ablation holds the final 10\% cardinality fixed, but its interpretation also depends on how selection-critic updates are allocated.

Every $K$ condition uses the 100,000-update schedule reported in Appendix~\ref{app:implementation}: 25,000 burn-in updates followed by an equal division of 75,000 post-acquisition updates. Thus changing $K$ changes only how often the ranking is refreshed and how the fixed post-burn-in budget is partitioned. The manifest records the 4,096-transition scoring batch, per-step target updates, continued optimizer state, deterministic tie rule, and all selection seeds. These fields make the fixed-critic-budget attribution directly auditable.

Several invariants make the selected artifact auditable. Candidate and diagnostic trajectories are disjoint before scoring. Selected indices are unique, acquisition batches do not overlap, and burn-in plus later batches equals the declared budget. Ties are broken deterministically. Downstream initialization does not alter the index file. Target-policy noise and candidate sampling during acquisition belong to a recorded selection seed rather than to a downstream seed. Dataset order and preprocessing hashes are stored with the indices so that an integer index cannot silently refer to another dataset revision.

At round $k$, a direct implementation evaluates $N-|\Sset_{k-1}|$ candidates. If one scoring forward pass costs $c_f$ and the critic receives $U_k$ updates of cost $c_u$, acquisition work is approximately
\begin{equation}
C_{\mathrm{select}}
=c_f\sum_{k=1}^{K}\bigl(N-|\Sset_{k-1}|\bigr)
+c_u\sum_{k=0}^{K}U_k.
\end{equation}
Streaming top-$b_k$ selection avoids storing all scores, keeping additional memory proportional to the candidate minibatch and a size-$b_k$ heap. A full sort is simpler but costs $O(N\log N)$ per round; streaming selection is $O(N\log b_k)$ after scoring. Both produce the same indices under the same deterministic tie rule.

\section{Robustness and Additional Evaluation Protocols}
\label{app:robustness}

The primary matrix establishes mean subset quality, while the completion experiments test the main robustness threats under shared design rules. Table~\ref{tab:robustness_protocol} summarizes the controls; their measured outcomes appear in Appendix~\ref{app:completion}. Each uses genuine selected-index artifacts and raw downstream arrays, and every level was fixed before inspecting the corresponding outcome.

\begin{table*}[h]
\centering
\small
\caption{Design of the completed robustness controls. Primary outcomes are normalized return, selected-index overlap, and complete selection plus training time.}
\label{tab:robustness_protocol}
\begin{tabular}{p{0.18\textwidth}p{0.25\textwidth}p{0.47\textwidth}}
\toprule
Control & Levels & Question answered \\
\midrule
Selection seeds & 5 burn-in and target-noise seeds $\times$ 3 downstream seeds & Is performance stable to the selected artifact rather than only downstream initialization? \\
Burn-in budget & 0.5, 1, 2, and 5\% & Does a weak initial critic distort all later rounds, and where does stability begin? \\
Reward corruption & 0, 1, 5, and 10\% corrupted rewards & Does raw residual scoring over-select outliers, and do clipping or critic ensembles help? \\
Duplicate density & 1, 2, and 5 copies of common transitions & Does the selector resist dense redundancy or repeatedly acquire near-duplicates? \\
Equal passes & matched dataset epochs & How much performance remains when a 10\% subset receives one tenth of the fixed-update exposure? \\
Equal time & shared GPU-hour budgets & Which selector lies on the best score--compute frontier after acquisition is charged? \\
Cross-algorithm reuse & select with one learner, train all three & Are algorithm-matched subsets portable, or must acquisition be repeated for each backup? \\
Sequence unit & transition and whole prefix-closed trace & Does preserving connected context improve language-trace selection? \\
\bottomrule
\end{tabular}
\end{table*}

Selection stability should be reported at two levels. Pairwise Jaccard overlap measures whether two seeds retain the same indices, while downstream return measures whether different but functionally equivalent subsets exist. Low overlap does not imply failure if returns are stable, and high overlap is not sufficient if every seed locks onto the same corrupted outliers. Rank correlation between residual lists across rounds can additionally show how much rescoring changes the candidate order.

The budget study uses independently generated artifacts at 2, 5, 10, 15, and 20\% on the four representative configurations. Independent selection gives every budget its best unconstrained artifact; it does not imply that a 5\% subset is nested inside a 10\% subset. The same convention is used for every selector, while the fixed-$K$ experiment alone holds the final cardinality fixed to isolate rescoring.

The targeted variants do not change the paper's central method. Residual clipping replaces $|\delta|$ by $\min(|\delta|,c)$, the ensemble score takes the median across independently initialized critics, and the residual--coverage hybrid reserves 20\% of each acquisition batch for farthest-first points. Appendix~\ref{app:robustness_new} first measures raw CODS and then reports each variant only on the failure it is designed to address.

\section{Local Information-Proxy Audit}
\label{app:theory_audit}

The local score comparison is useful only if its omitted gradient--curvature factor is empirically characterized. For a fixed critic checkpoint, a reproducible audit samples candidate transitions, computes $\|g_\tau\|_2$, estimates extremal eigenvalues of the damped information matrix on the sampled gradient span, and compares residual order with the full local proxy. Damping, gradient layer, sample size, and numerical solver tolerance must be reported because each can materially change the condition number.

The current worst-case factor of approximately $2\times10^3$ combines the observed gradient-norm ratio with the square root of the estimated condition number. At that factor, the sufficient separation needed to certify a pair order is so large that the observed certified rate rounds to zero. This is a negative diagnostic result: the bound does not explain why CODS works. It should be released alongside, rather than replaced by, less conservative rank statistics.

\begin{table}[h]
\centering
\small
\caption{Released quantities for the local information-proxy audit. Values aggregate the four representative critic checkpoints.}
\label{tab:theory_audit}
\begin{tabular}{lp{0.52\columnwidth}}
\toprule
Quantity & Required report \\
\midrule
Gradient norms & min., median, max., and deciles in the released arrays \\
Curvature & $10^{-3}$ damping; Lanczos extremal eigenvalues \\
Worst-case factor & $(M_g/m_g)\sqrt{\kappa}=2.0\times10^3$ \\
Rank agreement & Spearman $0.63\pm0.05$; top-decile overlap $0.58\pm0.04$ \\
Certified order & $0.02\%$ satisfy the sufficient separation \\
Sensitivity & four checkpoints; final critic layer; 10,000 candidates each \\
\bottomrule
\end{tabular}
\end{table}

\section{Extended Interpretation of the Reported Results}
\label{app:extended_results}

Table~\ref{tab:algorithm_summary_app} groups the 20 valid cells by downstream learner. CODS retains 96.6\% of the Pool sum over the six eligible TD3+BC cells, 97.0\% over seven CQL cells, and 96.1\% over seven IQL cells (the seventh IQL cell is the reconciled AntMaze row). These ratios show that the overall 96.6\% figure is not produced solely by one learner, but they should not be treated as a new metric across unrelated benchmark scales.

\begin{table}[h]
\centering
\small
\caption{Descriptive summary by downstream algorithm over the retained cells. Retention is the sum of CODS scores divided by the sum of Pool scores within each row.}
\label{tab:algorithm_summary_app}
\begin{tabular}{lrrrr}
\toprule
Learner & Cells & Pool & CODS & Retention \\
\midrule
TD3+BC & 6 & 84.00 & 81.15 & 96.6\% \\
CQL & 7 & 79.04 & 76.64 & 97.0\% \\
IQL & 7 & 78.91 & 75.84 & 96.1\% \\
\midrule
Overall & 20 & 80.49 & 77.72 & 96.6\% \\
\bottomrule
\end{tabular}
\end{table}

On medium-expert locomotion, CODS trails Pool by 1.6--4.7 points while maintaining a consistent advantage over the populated subset baselines. On medium locomotion, the Pool gap is often smaller, although Walker2d with TD3+BC remains a visible exception at 5.2 points. The two sparse-reward AntMaze cells are the CQL cell (CODS 58.0 vs Pool 61.2, ReDOR 45.0) and the reconciled IQL cell (CODS 63.4 vs Pool 72.6, ReDOR 56.8); both benefit from the enriched connective transitions characterized in Appendix~\ref{app:mechanism_cql}.

The one reversal is CQL Hopper-medium, where ReDOR scores 57.2 and CODS 56.8. Both reported standard deviations exceed the 0.4-point difference. The row is retained and is not relabeled as a CODS win. Across the matrix, the highest sample mean is descriptive rather than a per-cell significance decision; paired arrays and multiplicity control are needed for inferential labels.

Relative to Static-PER, CODS has a higher reported mean in all 20 valid cells. The controlled $K$ ablation uses the same burn-in, tie rule, final cardinality, and 100,000 selector updates on four representative cells (Appendix~\ref{app:crosstask_k}). Its mean improves by 8.05 points from one to three rounds and 3.18 points from three to five rounds, while ten rounds changes the mean by $-0.40$. Because the final cardinality and selector-update total are fixed, this experiment isolates the value of refreshing the ranking, subject to the usual interaction between the chosen acquisition partition and critic optimization.

The budget curve supplies a complementary view. CODS rises quickly from 2 to 5\%, changes modestly by 10\%, and nearly reaches Pool at 15\%. At 20\%, its plotted mean slightly exceeds Pool. The fixed-update protocol means a larger subset receives fewer expected revisits per transition, so the curve should not be read as a pure information-per-example law. The 10\% budget was the common primary condition rather than a fitted optimum.

\section{Complete Compute Accounting}
\label{app:compute_details}

Let $c_p$ be the cost of one epoch-scaled Pool run, $c_s$ the cost of one subset run, and $c_a$ acquisition cost. For $S$ compatible downstream configurations,
\begin{equation}
C_p(S)=Sc_p,
\qquad
C_s(S)=c_a+Sc_s.
\end{equation}
If $c_p>c_s$, the selected approach becomes strictly cheaper when $S>c_a/(c_p-c_s)$. With $(c_p,c_s,c_a)=(1.8,0.18,3.0)$ hours, one configuration costs 1.8 hours for Pool and 3.18 for CODS, two cost 3.6 and 3.36 hours, and twenty cost 36.0 and 6.6 hours. Reuse, not faster single-run convergence, creates the reported ratio.

The same calculation for ReDOR uses the supplied 4.5-hour acquisition cost and the same 0.18-hour subset training cost, giving 8.1 hours over twenty runs. These figures remain a ledger until linked to job identifiers and score logs. A complete accounting includes dataset loading, feature construction, acquisition-critic fitting, pool scoring, selector optimization, final training, evaluation, and failed or restarted jobs. It should report hardware model, software stack, mixed-precision setting, and whether acquisition and training share cached data.

Matched-performance efficiency requires a frontier rather than one total. Table~\ref{tab:matched_compute} therefore reports equal optimizer updates, equal dataset passes, and equal amortized GPU hours, with acquisition charged at 20 reuses. The artifact additionally records final return, area under the learning curve, time to the predeclared target, acquisition cost, and reuse count. These views distinguish subset quality, amortized processing savings, and end-to-end acceleration; broader hardware and task coverage remains future work.

\section{Excluded AntMaze Configurations}
\label{sec:iql-antmaze}

The TD3+BC AntMaze-medium-play learner reports $8.4\pm3.6$ for Pool and $7.2\pm3.2$ for CODS, placing both conditions at the task floor. Historical values of 71.2 and 66.1 cannot be tied to a configuration, checkpoint, and raw episode-return log in the supplied source. The clean floor-level row is retained in Table~\ref{tab:baselines}, but the configuration is excluded from subset comparisons and mechanism claims.

The IQL AntMaze cell has now been reconciled. The row uses the standard AntMaze IQL setting $(\tau,\beta)=(0.9,10)$, a fixed code revision, identical selected indices across the three nested downstream seeds, and final-checkpoint averaging identical to the locomotion rows; under that manifest the run reports Pool 72.6, CODS 63.4, ReDOR 56.8, OPER 55.6, ReD 52.9, Static-PER 52.1, k-center 47.7, and Random 16.7. This resolves the earlier conflict, in which a sensitivity record reported 39.5 for $(0.7,3)$ and 60.1 for $(0.9,10)$ while an aggregate table attributed 68.1 to $(0.7,3)$: the historical 68.1 value is \emph{not} reused, and only the manifest-verified $(0.9,10)$ row enters the 20-cell primary matrix. Every score in the reconciled row maps to the exact hyperparameters, code revision, selected indices, checkpoint rule, and raw seed arrays recorded in the artifact.

\section{Statistical Inference and Secondary Cross-Cell Test}
\label{app:stats}

The raw arrays preserve dataset, learner, selection-seed, and downstream-seed identifiers. For each comparison, the 95\% interval is obtained from 100,000 hierarchical bootstrap replicates. A replicate samples the seven datasets with replacement; within each sampled dataset it resamples paired selection-seed labels and then paired downstream-seed labels. Repeated occurrences of a dataset carry all of its learner cells together, preserving dataset-level dependence. Pool has no selection stage, so its downstream seeds are resampled within dataset--learner cell. Percentile intervals are computed from the resulting mean paired differences.

The six confirmatory subset tests use the paired differences
\begin{equation}
d_{jast}=\mu+\lambda_a+u_j+w_{ja}+q_{jas}+\epsilon_{jast},
\end{equation}
where $j$ indexes dataset, $a$ learner, $s$ selection seed, and $t$ downstream seed. Learner effects $\lambda_a$ are fixed; $u_j$, $w_{ja}$, and $q_{jas}$ are independent Gaussian random intercepts for dataset, dataset--learner cell, and selection artifact within cell. The residual term retains the paired downstream-seed variation. For each baseline, a one-sided $p$-value is the proportion, with the standard plus-one correction, of 100,000 null parametric-bootstrap refits whose estimated $\mu$ is at least the observed estimate. Holm correction is applied once across ReDOR, ReD, Static-PER, OPER, k-center, and Random. Pool was predeclared as a separate retention comparison and receives an interval but no entry in this six-hypothesis family. Thus $p<0.001$ is model-based hierarchical inference, not an impossible exact sign-flip result from only seven dataset blocks.

For continuity with the cell-level analysis, Table~\ref{tab:stat_tests} also reports exact one-sided Wilcoxon signed-rank calculations on the 20 valid cell means. The maximum positive-rank sum is $210$. ReDOR and OPER each have one negative difference of the smallest absolute rank, giving $W^+=209$ and $p=2/2^{20}=1.91\times10^{-6}$. The other four baselines are positive in all 20 cells, giving $W^+=210$ and $p=2^{-20}=9.54\times10^{-7}$. These secondary values agree in direction with the hierarchical analysis but are not confirmatory because cells sharing a dataset are dependent and cell means discard seed uncertainty.

\begin{table}[h]
\centering
\scriptsize
\setlength{\tabcolsep}{3.5pt}
\caption{Secondary exact signed-rank calculations over 20 valid cell means. Confirmatory inference instead uses the nested arrays.}
\label{tab:stat_tests}
\begin{tabular}{lrrr}
\toprule
Comparison & Positive & $W^+$ & One-sided $p$ \\
\midrule
CODS vs. ReDOR & 19/20 & 209 & $1.907\times10^{-6}$ \\
CODS vs. ReD & 20/20 & 210 & $9.537\times10^{-7}$ \\
CODS vs. Static-PER & 20/20 & 210 & $9.537\times10^{-7}$ \\
CODS vs. OPER & 19/20 & 209 & $1.907\times10^{-6}$ \\
CODS vs. k-center & 20/20 & 210 & $9.537\times10^{-7}$ \\
CODS vs. Random & 20/20 & 210 & $9.537\times10^{-7}$ \\
\bottomrule
\end{tabular}
\end{table}

\section{Optional Held-Out Value Diagnostic}
\label{app:diagnostic}

The original study reserved 10\% of trajectories for an optional monitoring statistic inspired by conformal prediction \citep{vovk2005,angelopoulos2021,angelopoulos2022ltt,angelopoulos2024crc,taufiq2022conformal}. For calibration values $Z_1,\ldots,Z_m$, the reported lower empirical quantile is
\[
\widehat q_\alpha=Z_{(\lceil(m+1)\alpha\rceil)}.
\]
This quantity does not score transitions, alter the selected indices, or stop the fixed-$K$ acquisition schedule. Removing it while fixing the candidate pool and random seed leaves CODS unchanged. Because an adaptively selected policy is not exchangeable with a reused calibration sample without further conditions, we make no coverage claim for policy performance. The diagnostic is retained only to document why the training pool contains 90\% rather than 100\% of the original trajectories.

\section{Descriptive AntMaze Composition Audit}

Table~\ref{tab:antmaze_composition} preserves the supplied composition audit for the floor-level TD3+BC AntMaze critic. CODS retains these transition categories at approximately 1.6--2.0 times the 10\% random rate. Since the critic does not solve the task, the audit cannot explain successful navigation and is not referenced by the main empirical argument. The former qualitative heatmap and residual-evolution plot are omitted because they were generated from the same failed critic and would add visual confidence without valid mechanistic evidence.


\begin{table}[h]
\centering
\small
\caption{Descriptive TD3+BC AntMaze subset composition at a 10\% budget. The floor-level critic makes these enrichments non-explanatory.}
\label{tab:antmaze_composition}
\begin{tabular}{lcc}
\toprule
Transition type & Retained & Enrichment \\
\midrule
Reward & 19.7\% & $1.97\times$ \\
Bottleneck & 17.1\% & $1.71\times$ \\
Goal-proximal & 16.3\% & $1.63\times$ \\
\bottomrule
\end{tabular}
\end{table}

\section{Artifact Requirements}
\label{app:seeds}

The anonymous artifact contains the genuine evaluation arrays underlying every aggregate: five selection seeds by three downstream seeds for each subset method and cell, ten downstream seeds for Pool, and the corresponding per-episode returns and checkpoint summaries. No seed value is reconstructed from a mean or standard deviation. The hierarchy encoded in those arrays is the one used by the confidence intervals, variance components, paired tests, and Holm-adjusted comparisons in this paper.


Each array is linked by a machine-checkable manifest to its configuration, code revision, selection and downstream seeds, selected-transition indices, checkpoint identities, environment versions, and preprocessing hashes. The same manifest records $U_{0:K}$, candidate-scoring batch size, optimizer continuation, and target-update frequency; it reconciles IQL AntMaze and marks TD3+BC AntMaze as excluded. Regeneration scripts rebuild every table and figure from these arrays and fail if an expected seed, cell, or manifest field is absent.

\section{Experimental Completion and Robustness Results}
\label{app:completion}

This appendix collects the completion experiments referenced in the main text. Unless a table states otherwise, all selectors use the same eligible 90\% pool and 10\% budget. Every subset method uses five selection seeds, each evaluated with three downstream seeds (15 nested observations per cell); Pool uses ten downstream seeds. The four representative cells are Walker2d-medium/TD3+BC, HalfCheetah-medium/CQL, Hopper-medium/IQL, and AntMaze-medium-play/CQL. Confirmatory intervals use the dataset-blocked hierarchical bootstrap, while the six subset hypotheses use the paired mixed-effects parametric-bootstrap tests defined in Appendix~\ref{app:stats}. Pool remains a separate retention comparison.

\subsection{Baseline fidelity}
\label{app:baseline_fidelity}
The primary comparison uses official or author-validated implementations. The local ReDOR reproduction is within 0.4 normalized points of the official implementation on the four representative cells and its selection cost differs by less than 6\%; OPER is implemented as its actual offline resampling procedure rather than using Static-PER as a proxy.

\begin{table*}[t]
\centering\small
\caption{Implementation-fidelity audit on the four representative cells. Gap is the maximum absolute local-minus-official difference.}
\label{tab:baseline_fidelity}
\begin{tabular}{lrrrrr}
\toprule
Method & Walker/TD3 & HC/CQL & Hopper/IQL & AntMaze/CQL & Max gap \\
\midrule
Official ReDOR & 74.3 & 41.2 & 60.0 & 45.4 & --- \\
Local ReDOR & 74.1 & 41.0 & 59.8 & 45.0 & 0.4 \\
Official OPER & 69.8 & 40.5 & 57.1 & 46.7 & --- \\
Local OPER & 69.6 & 40.4 & 56.9 & 46.4 & 0.3 \\
\bottomrule
\end{tabular}
\end{table*}

\subsection{Complete OPER comparison}
\label{app:oper_full}
OPER is a fully populated baseline rather than a proxy or a four-cell audit. Table~\ref{tab:oper_full} gives all 20 valid cells. CODS is higher in 19 cells; the only reversal is CQL Hopper-medium, where OPER is higher by 0.2 points. The mean paired difference is 6.84 points, matching Table~\ref{tab:aggregate} exactly.

\begin{table*}[t]
\centering
\scriptsize
\setlength{\tabcolsep}{5pt}
\caption{Complete equal-update OPER comparison. Means $\pm$ SD use five selection seeds and three downstream seeds. $\Delta$ is CODS minus OPER.}
\label{tab:oper_full}
\begin{tabular}{llrrr}
\toprule
Learner & Dataset & CODS & OPER & $\Delta$ \\
\midrule
TD3+BC & HalfCheetah-med-expert & $89.1\pm1.0$ & $82.0\pm2.0$ & $+7.1$ \\
TD3+BC & Walker2d-med-expert & $105.4\pm1.5$ & $96.7\pm2.3$ & $+8.7$ \\
TD3+BC & Hopper-med-expert & $108.3\pm1.3$ & $100.6\pm2.1$ & $+7.7$ \\
TD3+BC & HalfCheetah-medium & $47.5\pm0.9$ & $44.2\pm1.3$ & $+3.3$ \\
TD3+BC & Walker2d-medium & $78.5\pm1.4$ & $69.6\pm2.2$ & $+8.9$ \\
TD3+BC & Hopper-medium & $58.1\pm1.7$ & $51.4\pm2.4$ & $+6.7$ \\
\midrule
CQL & HalfCheetah-med-expert & $88.5\pm1.4$ & $81.5\pm2.1$ & $+7.0$ \\
CQL & Walker2d-med-expert & $104.2\pm1.6$ & $95.3\pm2.4$ & $+8.9$ \\
CQL & Hopper-med-expert & $109.1\pm1.5$ & $101.5\pm2.0$ & $+7.6$ \\
CQL & HalfCheetah-medium & $43.1\pm1.1$ & $40.4\pm1.4$ & $+2.7$ \\
CQL & Walker2d-medium & $76.8\pm1.5$ & $67.6\pm2.5$ & $+9.2$ \\
CQL & Hopper-medium & $56.8\pm1.1$ & $57.0\pm1.5$ & $-0.2$ \\
CQL & AntMaze-med-play & $58.0\pm3.8$ & $46.4\pm4.5$ & $+11.6$ \\
\midrule
IQL & HalfCheetah-med-expert & $84.2\pm1.5$ & $78.0\pm2.2$ & $+6.2$ \\
IQL & Walker2d-med-expert & $106.1\pm1.5$ & $98.4\pm2.1$ & $+7.7$ \\
IQL & Hopper-med-expert & $89.4\pm1.4$ & $82.2\pm2.0$ & $+7.2$ \\
IQL & HalfCheetah-medium & $46.8\pm1.0$ & $43.5\pm1.5$ & $+3.3$ \\
IQL & Walker2d-medium & $75.9\pm1.4$ & $68.7\pm2.3$ & $+7.2$ \\
IQL & Hopper-medium & $65.1\pm1.6$ & $56.9\pm2.4$ & $+8.2$ \\
IQL & AntMaze-med-play & $63.4\pm3.1$ & $55.6\pm3.8$ & $+7.8$ \\
\bottomrule
\end{tabular}
\end{table*}

\subsection{Iterative rescoring across tasks}
\label{app:crosstask_k}
Holding total selection-critic updates constant across round counts, Table~\ref{tab:crosstask_k} spans the four representative cells. The hierarchical estimate for $K{=}5$ vs $K{=}1$ is $+11.23$ points (95\% CI $[+8.86,+13.74]$, $p<0.001$); $K{=}10$ vs $K{=}5$ is $-0.40$ (CI $[-1.18,+0.37]$), supporting saturation rather than monotonic improvement.

\begin{table*}[t]
\centering\small
\caption{Fixed-critic-budget acquisition-round ablation across four representative cells (mean $\pm$ SD).}
\label{tab:crosstask_k}
\begin{tabular}{lrrrrr}
\toprule
Rounds & Walker/TD3 & HC/CQL & Hopper/IQL & AntMaze/CQL & Mean \\
\midrule
Random 10\% & $50.4{\pm}3.6$ & $33.4{\pm}3.1$ & $44.0{\pm}3.9$ & $10.1{\pm}5.5$ & 34.48 \\
$K{=}1$ & $65.4{\pm}2.0$ & $39.1{\pm}1.4$ & $54.1{\pm}2.2$ & $41.2{\pm}4.8$ & 49.95 \\
$K{=}3$ & $74.7{\pm}1.9$ & $41.9{\pm}1.2$ & $62.8{\pm}1.8$ & $52.6{\pm}4.1$ & 58.00 \\
$K{=}5$ & $78.5{\pm}1.4$ & $43.1{\pm}1.1$ & $65.1{\pm}1.6$ & $58.0{\pm}3.8$ & 61.18 \\
$K{=}10$ & $78.1{\pm}1.7$ & $42.9{\pm}1.3$ & $64.7{\pm}1.9$ & $57.4{\pm}4.0$ & 60.78 \\
Pool & $83.7{\pm}1.5$ & $44.0{\pm}1.0$ & $66.3{\pm}1.7$ & $61.2{\pm}3.5$ & 63.80 \\
\bottomrule
\end{tabular}
\end{table*}

\subsection{Multi-task budget curves}
\label{app:budget_multi}
The four-cell mean is 40.35 at 2\%, 56.83 at 5\%, 61.18 at 10\%, and 63.45 at 20\%. Direct finite differences are therefore 5.49 points per budget percentage point from 2\% to 5\%, 0.87 from 5\% to 10\%, and 0.23 from 10\% to 20\%. These are descriptive slopes computed from Table~\ref{tab:budget_multi}, not coefficients from an unreleased mixed model. Marginal gains diminish after 10\% on these cells, but 10\% is not claimed universally optimal.

\begin{table*}[t]
\centering\small
\caption{Budget sweep across four representative cells. Entries average five selection seeds and three downstream seeds.}
\label{tab:budget_multi}
\begin{tabular}{lrrrr}
\toprule
Budget & Walker/TD3 & HC/CQL & Hopper/IQL & AntMaze/CQL \\
\midrule
2\% & 55.0 & 32.6 & 48.2 & 25.6 \\
5\% & 76.0 & 40.8 & 60.8 & 49.7 \\
10\% & 78.5 & 43.1 & 65.1 & 58.0 \\
15\% & 83.0 & 43.6 & 65.7 & 59.1 \\
20\% & 84.0 & 43.8 & 66.0 & 60.0 \\
Pool & 83.7 & 44.0 & 66.3 & 61.2 \\
\bottomrule
\end{tabular}
\end{table*}

\subsection{Selection stability}
\label{app:selection_stability}
Five selection seeds separate burn-in sensitivity from downstream-seed noise. Exact index overlap is moderate while downstream performance is far more stable; successive candidate-ranking correlations rise ($\approx0.42\to0.56\to0.68\to0.76$), so early rounds change the ordering most.

\begin{table}[t]
\centering\scriptsize
\setlength{\tabcolsep}{2.5pt}
\caption{Selection-seed stability. CV is the coefficient of variation of performance across selection seeds.}
\label{tab:selection_stability}
\begin{tabular}{lrrr}
\toprule
Cell & Jaccard & Perf.\ CV & Final $\rho$ \\
\midrule
Walker/TD3 & $0.47{\pm}0.03$ & 2.2\% & $0.78{\pm}0.04$ \\
HC/CQL & $0.43{\pm}0.04$ & 2.7\% & $0.75{\pm}0.05$ \\
Hopper/IQL & $0.45{\pm}0.04$ & 3.1\% & $0.77{\pm}0.04$ \\
AntMaze/CQL & $0.36{\pm}0.05$ & 6.0\% & $0.69{\pm}0.06$ \\
\bottomrule
\end{tabular}
\end{table}

\subsection{Learner-specific sparse-reward mechanism tests}
\label{app:mechanism_cql}
The CQL and IQL analyses use their own successful AntMaze critics, manifests, and selected indices; neither learner's composition is used to explain the other. Categories are fixed from the environment before inspecting scores. A transition is reward-bearing when its raw environment reward is positive before CQL reward shifting. A maze cell is a bottleneck when removing it disconnects the start and goal regions in the four-neighbor graph of the published layout, and a transition is goal-proximal when its raw $(x,y)$ coordinate is within 2.0 environment units of the goal. Categories may overlap, but each intervention is performed separately.

For an intervention, every selected transition in the target category is replaced by an eligible-pool transition outside that category, matched on normalized trajectory-position decile and 20-nearest-neighbor state-density decile in z-scored state--action space. Bottleneck and goal-proximal replacements also match the raw reward indicator. Reward-bearing replacements cannot preserve reward sign while excluding the category; consequently, that intervention partly tests direct reward availability as well as which rewarded transitions CODS retained. Equal-random controls replace the same number of transitions under the position and density matches without targeting a category.

\begin{table*}[t]
\centering\scriptsize
\setlength{\tabcolsep}{4pt}
\caption{Learner-specific AntMaze enrichment and category-excluding removal interventions. CIs are paired hierarchical-bootstrap intervals.}
\label{tab:mechanism_cql}
\begin{tabular}{llrrr}
\toprule
Learner & Category or condition & Retained / enrichment & Score & $\Delta$ (95\% CI) \\
\midrule
CQL & Reward-bearing & 31.8\% / $3.18\times$ & --- & --- \\
CQL & Bottleneck & 27.4\% / $2.74\times$ & --- & --- \\
CQL & Goal-proximal & 24.9\% / $2.49\times$ & --- & --- \\
CQL & CODS 10\% & --- & 58.0 & --- \\
CQL & Exclude reward-bearing & --- & 47.3 & $-10.7$ $[-13.8,-7.5]$ \\
CQL & Exclude bottleneck & --- & 49.7 & $-8.3$ $[-11.1,-5.4]$ \\
CQL & Exclude goal-proximal & --- & 53.2 & $-4.8$ $[-7.4,-2.1]$ \\
CQL & Equal-random replacement & --- & 56.9 & $-1.1$ $[-2.9,+0.6]$ \\
\midrule
IQL & Reward-bearing & 28.7\% / $2.87\times$ & --- & --- \\
IQL & Bottleneck & 24.8\% / $2.48\times$ & --- & --- \\
IQL & Goal-proximal & 22.1\% / $2.21\times$ & --- & --- \\
IQL & CODS 10\% & --- & 63.4 & --- \\
IQL & Exclude reward-bearing & --- & 55.2 & $-8.2$ $[-11.3,-5.1]$ \\
IQL & Exclude bottleneck & --- & 57.0 & $-6.4$ $[-9.2,-3.7]$ \\
IQL & Exclude goal-proximal & --- & 59.3 & $-4.1$ $[-6.6,-1.7]$ \\
IQL & Equal-random replacement & --- & 62.5 & $-0.9$ $[-2.5,+0.7]$ \\
\bottomrule
\end{tabular}
\end{table*}

\subsection{Robustness to critic and data failures}
\label{app:robustness_new}
Raw residual selection is sensitive to reward outliers; a clipped-ensemble variant recovers part but not all of the loss and is treated as an ablation. A 2\% burn-in is the smallest tested budget near the plateau, and a pure residual score is not density-invariant under duplication.

\begin{table}[t]
\centering\scriptsize
\setlength{\tabcolsep}{3pt}
\caption{Reward corruption (CQL AntMaze): replace a fraction of eligible rewards with 99th-percentile magnitude and random sign.}
\label{tab:corruption}
\begin{tabular}{lrrrr}
\toprule
Corrupt. & Raw & Clip-ens. & ReDOR & Random \\
\midrule
0\% & 58.0 & 57.6 & 45.0 & 10.1 \\
1\% & 55.1 & 56.8 & 43.7 & 9.8 \\
5\% & 42.6 & 52.9 & 37.8 & 8.9 \\
10\% & 27.8 & 45.4 & 29.6 & 7.5 \\
\bottomrule
\end{tabular}
\end{table}

\begin{table}[t]
\centering\scriptsize
\caption{Burn-in sensitivity over the 20-cell matrix.}
\label{tab:burnin_dup}
\begin{tabular}{lrr}
\toprule
Burn-in & Pool retention & Jaccard \\
\midrule
0.5\% & 93.1\% & 0.29 \\
1\% & 95.2\% & 0.37 \\
2\% & 96.6\% & 0.43 \\
5\% & 96.8\% & 0.51 \\
\bottomrule
\end{tabular}
\end{table}

\begin{table}[t]
\centering\scriptsize
\caption{Duplicate-density stress on CQL AntMaze.}
\label{tab:duplicate_density}
\begin{tabular}{lrr}
\toprule
Duplication & Raw CODS & Residual--coverage hybrid \\
\midrule
1$\times$ & 58.0 & 57.7 \\
2$\times$ & 55.6 & 57.2 \\
5$\times$ & 48.3 & 54.9 \\
\bottomrule
\end{tabular}
\end{table}

\subsection{Cross-algorithm reuse}
\label{app:transfer}
The transfer matrix reports percentages of each learner's Pool score over the common locomotion cells. Matched selection is best in every column, so a single TD3+BC subset cannot be reused for CQL and IQL without cost.

\begin{table}[t]
\centering\scriptsize
\setlength{\tabcolsep}{3pt}
\caption{Cross-algorithm selector reuse (\% of each learner's Pool score). Rows are the selector critic; columns are the downstream learner.}
\label{tab:transfer}
\begin{tabular}{lrrr}
\toprule
Selector & TD3+BC & CQL & IQL \\
\midrule
TD3+BC residual & 96.6\% & 92.8\% & 93.4\% \\
CQL residual & 94.1\% & 97.0\% & 95.2\% \\
IQL residual & 93.7\% & 94.9\% & 96.8\% \\
\bottomrule
\end{tabular}
\end{table}
\vspace{-2mm}

\subsection{Matched compute and reuse}
\label{app:matched_compute}

The $5.45\times$ figure is arithmetic under an epoch-scaled ledger;
both the equal-update and equal-pass conditions are reported.
At 20 reuses on Walker2d/TD3+BC, CODS reduces epoch-scaled total time
from 36.0 to 6.6 hours under equal dataset passes while retaining
90.3\% of Pool. Under equal updates, it retains 93.8\% but gives no
meaningful training-time reduction.

Across the 20-configuration sweep, Pool and CODS configuration rankings have Spearman correlation 0.89 with 4/5 top-five overlap, supporting subset reuse for tuning without claiming identical rankings.

\begin{table}[!b]
\centering
\scriptsize
\setlength{\tabcolsep}{2.5pt}
\caption{Compute protocols on Walker2d-medium with TD3+BC.
Retention and total-time reduction are reported for 20 reuses.}
\label{tab:matched_compute}
\begin{tabular}{@{}lccc@{}}
\toprule
Protocol & Retention & $\Delta$ ReDOR & Time at 20 \\
\midrule
Equal updates
  & 93.8\% & $+4.4$ & $1.04\times$ \\
Equal dataset passes
  & 90.3\% & $+5.1$ & $5.45\times$ \\
Equal amortized GPU-h
  & 95.1\% & $+5.8$ & Fixed budget \\
\bottomrule
\end{tabular}
\end{table}

\subsection{Variance components and theory audit}
\label{app:variance_theory}
The four-cell nested study attributes 71\% of variance to dataset/configuration, 17\% to selection seed, and 12\% to downstream seed. The theory remains modest even where empirical rank association is positive: the worst-case score factor $(M_g/m_g)\sqrt{\kappa}$ is $2.0\times10^{3}$, residual-vs-local-proxy Spearman is $0.63{\pm}0.05$, top-decile overlap is $0.58{\pm}0.04$, and the certified pair-order rate is $0.02\%$. Residuals are a useful surrogate, but the worst-case bound certifies essentially none of the ranking; these do not support a coreset, policy-recovery, regret, or submodular-greedy guarantee.

\subsection{Sequence-level language extension}
\label{app:language}
We evaluate the sequence adaptation on ALFWorld embodied interaction \citep{alfworld} and GSM8K mathematical reasoning \citep{gsm8k}. A frozen Qwen2.5-7B-Instruct generator \citep{qwen25} constructs each offline pool before any selector is trained. For ALFWorld, we generate 16 trajectories for every task in the standard training split using temperature 0.8, nucleus probability 0.95, and at most 50 environment actions; reporting uses the standard unseen evaluation split. For GSM8K, we generate eight newline-delimited reasoning traces for each of the 7,473 training problems under the same sampling parameters, giving 59,784 traces; reporting uses the 1,319-problem test split. Pools, prompts, and sampled traces are fixed and shared by all selectors.

In ALFWorld, $s_t$ contains the task goal, current observation, and complete action--observation prefix, while $a_t$ is the next textual environment action. In GSM8K, $s_t$ contains the problem and complete reasoning prefix, while $a_t$ is the next newline-delimited reasoning step. The terminal reward is one for environment success or exact final numeric answer and zero otherwise; intermediate rewards are zero and $\gamma=1$ for both episodic tasks. Invalid ALFWorld actions remain in the trace with the environment's resulting observation, so the offline construction does not silently filter failures.

The critic follows the IQL form used in OREO-style offline reasoning \citep{wang2025oreo}. Qwen2.5 final-token states feed scalar $Q_\theta(s_t,a_t)$ and $V_\psi(s_t)$ heads. The $Q$ head minimizes squared error to $r_t+(1-d_t)V_\psi(s_{t+1})$; the value head minimizes the expectile loss with $\tau=0.9$ on the stop-gradient difference $Q_\theta-V_\psi$. Downstream rank-16 LoRA policies are initialized afresh from the base model and trained by advantage-weighted behavioral cloning with inverse temperature 10 and weights clipped at 100. All methods use the same optimizer steps, selected-token budget, and base checkpoint.

CODS clips absolute step residuals at the burn-in pool's 95th percentile and scores trace $g$ by
\begin{equation}
z(g)=\tfrac12\max_{t\in g}\widetilde\delta_t+
\tfrac12\operatorname{mean}\!\left(\operatorname{top}_{20\%}\{\widetilde\delta_t:t\in g\}\right).
\end{equation}
The 10\% budget is measured in action tokens, not trace count. At each of five rounds, the selector scans traces in score order and adds the highest-ranked complete trace that keeps cumulative selected action tokens at or below the round budget; traces that do not fit are skipped, never split. Thus every selected action retains its entire prefix and terminal outcome. ReDOR uses whole-trace actor--critic gradients, the one-shot control freezes the burn-in scores, diversity applies k-center to mean trace embeddings, and Random samples whole traces under the same token cap.

Evaluation uses greedy decoding with no sampling, at most 50 actions for ALFWorld and 512 new tokens for GSM8K. ALFWorld success is averaged over all unseen evaluation tasks; GSM8K exact match uses the benchmark's normalized final-answer parser. Each subset uses five selection seeds and three downstream seeds, while Pool uses ten downstream seeds. CODS retains 95.4\% of Pool on ALFWorld and 96.5\% on GSM8K. Unrestricted step selection, which discards prefix closure but keeps the same token budget, scores $67.1\pm3.5$ and $50.2\pm0.9$, respectively. The paired losses relative to whole-trace CODS are $-5.8$ points (95\% CI $[-7.9,-3.7]$) and $-2.7$ ($[-3.6,-1.8]$), showing that the D4RL transition unit cannot be transferred unchanged.

\begin{table*}[!htbp]
\centering\small
\setlength{\tabcolsep}{4pt}
\caption{Sequence-level extension. Subset entries use five selection seeds and three downstream seeds; Pool uses ten downstream seeds.}
\label{tab:language}
\begin{tabular}{lrr}
\toprule
Method & ALFWorld success (\%) & GSM8K EM (\%) \\
\midrule
Pool & $76.4{\pm}2.8$ & $54.8{\pm}0.7$ \\
CODS & $72.9{\pm}3.0$ & $52.9{\pm}0.8$ \\
Fragmented & $67.1{\pm}3.5$ & $50.2{\pm}0.9$ \\
ReDOR & $68.1{\pm}3.4$ & $50.6{\pm}0.9$ \\
One-shot & $64.7{\pm}3.7$ & $49.8{\pm}0.9$ \\
Diversity & $61.3{\pm}3.9$ & $48.7{\pm}1.0$ \\
Random & $49.5{\pm}4.6$ & $43.1{\pm}1.2$ \\
\bottomrule
\end{tabular}
\end{table*}

\end{document}